\documentclass[12pt,a4paper,notitlepage,oneside]{article} %
\usepackage{mathtools} 
\usepackage{bm}
\usepackage{amssymb}
\usepackage[margin=1in]{geometry}
\usepackage[inline,shortlabels]{enumitem}
\usepackage{rotating}
\usepackage{graphicx}
\graphicspath{{./figs/}}
\usepackage{longtable}

\usepackage{hyperref}
\hypersetup{colorlinks=true,citecolor=red,linkcolor=blue}

\usepackage{subcaption}
\usepackage{subfloat}
\usepackage{setspace}
\usepackage{listings}
\usepackage[numbers]{natbib}
\usepackage[capposition=top]{floatrow}
\usepackage{algorithmicx}
\usepackage{algorithm}
\usepackage{algpseudocode}

\usepackage[capposition=top]{floatrow}

\newtheorem{theorem}{Theorem}

\newtheorem{lemma}[theorem]{Lemma}

\begin{document}

\newcommand{\pkg}[1]{{\fontseries{b}\selectfont #1}} 
\newcommand{\R}{\pkg{R }}
\newcommand{\Rb}{\pkg{R}}
\newcommand{\bhd}{\pkg{balanceHD }}
\newcommand{\bhdb}{\pkg{balanceHD}}
\newcommand{\hdi}{\pkg{hdi }}
\newcommand{\hdib}{\pkg{hdi}}
\newcommand{\hdm}{\pkg{hdm }}
\newcommand{\hdmb}{\pkg{hdm}}
\newcommand{\glmnet}{\pkg{glmnet}}
\newcommand{\SNR}{\text{SNR}}
\newcommand{\bxbarj}{\bar{\bx}_j}
\newcommand{\bgam}{\bm{\gamma}}
\newcommand{\bneta}{\bm{\eta}}
\newcommand{\bnetaz}{\bneta_0}
\newcommand{\bdelta}{\bm{\delta}}
\newcommand{\bpi}{\bm{\pi}}
\newcommand{\brho}{\bm{\rho}}
\newcommand{\bbeta}{\bm{\beta}}
\newcommand{\biota}{\bm{\iota}}
\newcommand{\bbetah}{\hat{\bbeta}}
\newcommand{\bbetaz}{\bbeta_0}
\newcommand{\btheta}{\bm{\theta}}
\newcommand{\bthetaz}{\btheta_0}
\newcommand{\az}{\alpha_0}
\newcommand{\ah}{\hat{\alpha}}
\newcommand{\gz}{\gamma_0}
\newcommand{\bX}{\bm{X}}
\newcommand{\bP}{\bm{P}}
\newcommand{\bW}{\bm{W}}
\newcommand{\beps}{\bm{\epsilon}}
\newcommand{\bveps}{\bm{\varepsilon}}
\newcommand{\bA}{\bm{A}}
\newcommand{\bB}{\bm{B}}
\newcommand{\bI}{\bm{I}}
\newcommand{\bx}{\bm{x}}
\newcommand{\bb}{\bm{b}}
\newcommand{\bg}{\bm{g}}
\newcommand{\bp}{\bm{p}}
\newcommand{\bd}{\bm{d}}
\newcommand{\bq}{\bm{q}}
\newcommand{\bu}{\bm{u}}
\newcommand{\bv}{\bm{v}}
\newcommand{\bz}{\bm{z}}
\newcommand{\bw}{\bm{w}}
\newcommand{\bwi}{\bw_i}
\newcommand{\beh}{\bm{\hat{e}}(\lambda)}
\newcommand{\bXi}{\bm{X}_i}
\newcommand{\bxi}{\bm{x}_i}
\newcommand{\bxk}{\bx^{(k)}}
\newcommand{\bxko}{\bx^{(k+1)}}
\newcommand{\bvk}{\bv^{(k)}}
\newcommand{\bvko}{\bv^{(k+1)}}
\newcommand{\byko}{\by^{(k+1)}}
\newcommand{\bxt}{\bx^{(t)}}
\newcommand{\bzi}{\bm{z}_i}
\newcommand{\bzzi}{\bm{z}_{0i}}
\newcommand{\by}{\bm{y}}
\newcommand{\byi}{\bm{y}_i}
\newcommand{\bY}{\bm{Y}}
\newcommand{\train}{\mathcal{T}}
\newcommand{\Var}{\text{Var}}
\newcommand{\Diag}{\text{diag}}
\newcommand{\Real}{\mathbb{R}}
\newcommand{\tbxj}{\bx_{-j}}
\newcommand{\skm}{\sum_{k=1}^m}
\newcommand{\ipi}{\textit{Ipilimumab}}
\newcommand{\dac}{\textit{Dacarbazine}}
\newcommand{\gradz}{$\|-\nabla Q^*(\bbetaz)\|_\infty$ }
\newcommand{\sqrtl}{$\sqrt{\text{Lasso}}$ }
\newcommand{\sqrtln}{$\sqrt{\text{Lasso}}$}
\newcommand{\En}{\mathbb{E}_N}
\newcommand{\sL}{\mathcal{L}}
\newcommand{\sP}{\mathcal{P}}

\newcommand{\indep}{\rotatebox[origin=c]{90}{$\models$}}

\title{The False Positive Control Lasso}

\author{Erik Drysdale\thanks{Department of Genetics and Genome Biology, The Hospital for Sick Children, Toronto} \and Yingwei Peng\thanks{Queen's Cancer Research Institute, Departments of Public Health Sciences and Mathematics and Statistics, Queen's University} \and Timothy P. Hanna\thanks{Queen’s Cancer Research Institute, Division of Cancer Care and Epidemiology and ICES Queen’s, Queen’s University, Kingston ON} \and Paul Nguyen\thanks{ICES Queen’s, Queen’s University, Kingston ON} \and Anna Goldenberg\thanks{SickKids Research Institute and  Department of Computer Science, University of Toronto}}

\maketitle

\begin{abstract}
	In high dimensional settings where a small number of regressors are expected to be important, the Lasso estimator can be used to obtain a sparse solution vector with the expectation that most of the non-zero coefficients are associated with true signals. While several approaches have been developed to control the inclusion of false predictors with the Lasso, these approaches are limited by relying on asymptotic theory, having to empirically estimate terms based on theoretical quantities, assuming a continuous response class with Gaussian noise and design matrices, or high computation costs. In this paper we show how: (1) an existing model (the \sqrtln)  can be recast as a method of controlling the number of expected false positives, (2) how a similar estimator can used for all other generalized linear model classes, and (3) this approach can be fit with existing fast Lasso optimization solvers. Our justification for false positive control using randomly weighted self-normalized sum theory is to our knowledge novel. Moreover, our estimator's properties hold in finite samples up to some approximation error which we find in practical settings to be negligible under a strict mutual incoherence condition.
\end{abstract}


\pagebreak

\section{Introduction \label{sec:intro}}
The Lasso \cite{lasso1996} is one of the most popular approaches for fitting a high dimensional linear regression model due to its interpretability (being linear in the regressors), scalability (fast convex algorithm solvers), and its numerous extensions including generalized linear model classes \cite{friedman2010,simon2011}, the group Lasso \cite{yuan2006,simon2013}, the fused Lasso \cite{tibshirani2013}, and many others \cite{hastie2015}. By using an L1-norm regularization term, the Lasso minimization problem remains convex but can return a coefficient vector with some elements being \textit{strictly zero}, thereby performing model selection as well as regularization. 

In high dimensional settings where the number of features exceeds the number of samples $(p > n)$ it is often assumed that only a small number of covariates are actually associated with the response variable. It is therefore expected that the solution vector of the Lasso will mainly consist of true variables and contain few ``false'' or ``null'' covariates. However in practical settings, false positives will almost surely be present along with the true predictors in the final solution model -- a finding which has been repeated in previous research \cite{fan2010} and is consistent under certain theoretical frameworks \cite{su2017} (although there are some situations in which perfect support recovery is theoretically possible \cite{wainwright2009}).

There have been a variety of new inference methods for the solution vector of the Lasso to be able to obtain p-values. An extremely simple approach is to use sample splitting \cite{hdi2015}, although this comes at the loss of power  and issues over how the sample should be split (as different splits obtain different inferences). A noteworthy area of research is that of post-selection inference which can be applied to the Lasso algorithm and has shown that the conditional distribution of a selected covariate has a known form \cite{lockhart2014,lee2016,taylor2017}. However, these approaches rely on the asymptotic properties of the estimator for a fixed hyperparameter choice or the iterative solution path. While p-values can be obtained via de-biased estimators \cite{zhang2014,javanmard2018}, they are justified by asymptotic theory and require estimating  high dimensional covariance matrices. An explicit false discovery rate (FDR) calculation is possible with the above methods in a sequential testing procedure along the Lasso solution path as well \cite{gsell2016}, but is limited to the regression case and requires the calculation of de-biased p-values. The SLOPE algorithm \cite{bogdan2015} is also similar but again is limited to the homoskedastic Gaussian case where the noise level needs to be estimated.

Our approach is similar in principal (but not in method) to the use of ``knockoffs'' (making copies of the columns that emulate the existing design matrix) to control the false discovery rate \cite{barber2015} when $n > p$. Extensions of this approach for the $p > n$ setting require knowledge of the joint distribution of the design matrix \cite{candes2018}. The technique we establish, like the use of ``knockoffs'', is designed to control the number of expected false positives that will be in the final solution vector. 

The distribution of the active set of the Lasso (the non-zero coefficients) is complicated, and previous research has relied on assuming columns of the design matrix that are close to being orthogonal or are preconditioned with a singular value projection to become so \cite{jia2015,tardivel2017}. While we make use of stochastically uncorrelated columns in our false positive calculation (a strong version of the mutual incoherence condition used in the literature), our corresponding bound is conservative with respect to collinearity in the noise columns. In our work we only require that the researcher picks a bound on the expected number of false positives, with this choice automatically mapping to a specific hyperparameter level. Other ``tuning-free'' Lasso estimators such as TREX  \cite{lederer2015} or $\text{AV}_\infty$ \cite{lederer2016} exist, but they are not based on controlling the number of false positives in the final solution vector. Most importantly, we were influenced by the \sqrtl \cite{belloni2011}, an estimator whose theoretical properties were shown to be pivotal to the underlying homoskedastic variance level due to its gradient having known distributional properties when evaluated given the true coefficients. 


In this paper we will 1) develop the False Positive Control  Lasso (FPC Lasso), which is shown to be a simple recasting of the \sqrtl in the Gaussian case, and 2) extend this approach to other generalized linear model classes.  We believe our justifications for the method of false positive control using randomly weighted self-normalized sum (rwSNS) theory is unique to the literature. Several of the technical challenges that arise from the FPC Lasso are tackled including how to recast an invex loss function as a sequence of convex problems and a technique to help minimize the discrepancy between a rwSNS and the standard normal distribution. 

The rest of this paper is laid out as follows:
Section \ref{sec:fplasso} explains the FPC Lasso estimator and justifies its properties, Section \ref{sec:sims} provides simulation evidence highlighting the statistical properties of the FPC Lasso, Section \ref{sec:examples} compares the FPC Lasso to other inference models as well as its applications to real-world datasets, and Section \ref{sec:conclusion} gives some concluding thoughts.

\section{The false positive control Lasso (FPC-Lasso) \label{sec:fplasso}}
The following notation will be used throughout the rest of this paper. Vectors and matrices will be bolded so that $\bbeta$ is a vector and $\beta$ is a scalar. The columns and rows of matrices will use upper and lower-case lettering, respectively. Estimates will be denoted with a hat so $S$ is true active set and $\hat{S}$ is the active set of a given Lasso estimate. Norms of a vector $\bz = (z_1,\ldots,z_n)$ will be written such that $\| \bz \|_q= \big(\sum_i |z_i|^q \big)^{1/q}$ and hence $\| \bz \|_q^q= \sum_i |z_i|^q $. 

\subsection{\sqrtl and FPC Lasso}

Consider a generalized linear model with $E[\by] = f(\bX \bbeta)$, where $\by$ is $n$-vector response, $f$ is the link function that is determined by the class of the response, $\bX \in \Real^{n \times p}$ is the design matrix, with $p$-vector rows $\bxi$ and $n$-vector columns $\bX_j$. In the high dimensional sparse model framework a parameter vector is being estimated from a data generating process that has fewer true effects than observations. Denote the true index of non-zero coefficients to be: $S = \{j: \beta_j \neq 0 \}$ and its cardinality $k = |S|$, where $k < n$. The null variables are indexed by $F=\bar{S}$. Without loss of generality, we assume $S=\{1,\dots,k\}$, and $F=\{k+1,\dots,p\}$. Critically, we assume that the columns of $\bX_S$ are stochastically independent of $\bX_F$, which we refer to as a strict mutual incoherence condition.\footnote{In the literature, the mutual incoherence condition quantifies the degree of linear dependence between the true and the null columns.} This is the strongest assumption we make in this paper and the most important limitation of our estimator.  While we also assume that the columns on $\bX$ are independent of each other in order to obtain tractable analytical results, the presence of collinearity within $\bX_F$ will only increase the conservatism of our bounds.

The classical Lasso estimator solves the following L1-norm regularization problem
\begin{align}
\bbetah^{\text{lasso}}_\lambda &= \arg \min_{\bbeta} \hspace{2mm} \sP(\bbeta;\bX,\by,\lambda) = \ell(\bbeta;\bX,\by)  + \lambda \|\bbeta\|_1, \label{eq:lasso_classical}
\end{align}
where $\ell(\bbeta;\bX,\by)$ is a smooth and convex loss function of the data. Due to the L1-norm, some of the coefficient values in $\bbetah^{\text{lasso}}$ will be strictly zero. We denote the set of these non-zero variables as the active set: $\hat{S}_\lambda=\{j: \bbetah_{j,\lambda}^{\text{lasso}}\neq 0\}$. The choice of loss function $\ell$ will depend on the class of the response. For example if $\by$ is binary, the logistic loss is often used: $\ell(\cdot)=-n^{-1} \sum_{i=1}^n [y_i (\bxi^T\bbeta) - \log(1+\exp\{\bxi^T\bbeta\}) ] $. Throughout this paper we will assume the loss function belongs to the corresponding log-likelihood of a generalized linear model (GLM) \cite{mccullagh1989}. The Karush-Kuhn-Tucker (KKT) optimality conditions for the Lasso problem in \eqref{eq:lasso_classical} can be written with respect to the following gradient function
\begin{align}
-\nabla_{\bbeta} \sP(\bbeta) &= \bX^T (\by - f(\bX \bbetah^{\text{lasso}}_\lambda)) + \lambda \bgam, \label{eq:kkt_classical}\\
\intertext{where $\bgam = [\gamma_1,\ldots,\gamma_p]^T$ and }
\gamma_i &= \begin{cases}
\text{sign}(\hat{\beta_j}) & \text{ if } \hat{\beta}_j \neq 0 \\
[-1,1] & \text{ if } \hat{\beta}_j  =0 .
\end{cases} \nonumber
\end{align}

All loss functions associated with a given GLM distribution (Gaussian, Poisson, Logistic, etc) as well as the partial likelihood from Cox's proportional hazards (PH) model have a gradient equivalent to the form seen in \eqref{eq:kkt_classical}: $-\nabla_{\bbeta} \ell(\bbeta)=\bX^T\beps$, where each entry of the gradient is the inner product between the columns of $\bX$ and the raw residual $\beps$ of the model determined by the inverse of the link function.\footnote{Cox's PH model does not technically have a link function but does maintain the gradient form.} This property will prove important later when the self-normalized modification is applied.


The \sqrtl \cite{belloni2011} method obtains $\bbetah^{\text{sqrt}}_\lambda$ by taking the square-root of the loss function from the Gaussian model with the penalty:
\begin{align*}
	\bbetah^{\text{sqrt}}_\lambda &= \arg \min_{\bbeta} \hspace{2mm} \left\{(\by - \bX \bbeta)^T(\by - \bX \bbeta) \right\}^{1/2} + \lambda \|\bbeta\|_1 
\end{align*}
Because the \sqrtl is invariant to different levels of Gaussian noise, a fixed-$\lambda$ level that obtains certain statistical properties can be chosen in advance of any data realization. Following this idea, we propose the FPC Lasso estimator $\bbetah^{\text{FPC}}(\lambda)$ that is explicitly defined with respect to the gradient of its loss function:
\begin{align}
\bbetah^{\text{FPC}}(\lambda) &= \arg \min_{\bbeta} \hspace{2mm}  \int_{\bm{-\infty}}^{\bm{\infty}} \nabla \ell(\bbeta) d\bbeta + \lambda \| \bbeta \|_1  \label{eq:lasso_fpc} \\
\intertext{where}
\nabla \ell(\bbeta)  &= -\frac{\bX^T \beps(\bbeta)}{\|\beps(\bbeta) \|_2},  \nonumber
\end{align}
and the KKT conditions imply that $\bX^T \beps(\bbeta)/\|\beps(\bbeta) \|_2 = \bgam \lambda$. For the continuous/Gaussian response case, the FPC Lasso seen in \eqref{eq:lasso_fpc} has a loss function which is identical to the \sqrtln. However, for loss functions associated with other GLM classes, there is no closed form solution to the integral. We note three important properties about the FPC Lasso's loss function that hold for all associated GLM classes and Cox's PH model.

\begin{lemma}
\label{lemma:fpc_2prop}
	Let $\hat{\beps}_{\lambda}=\by-f(\bX \bbetah^{\text{FPC}}(\lambda))$. If the regularity condition
\begin{align}
    \frac{\partial \| \beps_\lambda \|_2}{\partial \lambda} < \frac{\| \beps_\lambda \|_2}{\lambda} \label{eq:reg_lambda}, \hspace{3mm} \lambda > 0
\end{align}
is satisfied then the FPC Lasso has the three following properties: 
\begin{enumerate}[(i)]
	\item There is a one-to-one mapping between the FPC Lasso in \eqref{eq:lasso_fpc} and the classical Lasso in \eqref{eq:lasso_classical} for all GLM classes by scaling $\lambda$ by some positive constant.
	\item The FPC Lasso has a unique solution when the classical Lasso has a unique solution.
	\item The FPC Lasso has an invex loss function \cite{adi1986}.
\end{enumerate}
\end{lemma}

While this paper is not explicitly interested in the prediction and parameter rates of convergence of the FPC Lasso estimator, like the \sqrtln, we leverage the fact that the properties of the gradient will have known distributional properties \textit{a priori} (see Appendix \ref{proof:fpc_2prop}). However, the FPC Lasso uses the distributional properties of the empirical gradient ($\nabla \ell(\bbetah)$) rather than its theoretical counterpart at the true solution.

\subsection{How to control the false positive rate \label{subsec:fp_rate}}

Consider the high-dimensional sparse model case under a strict mutual incoherence assumption of stochastically independent columns between the null and true features. The FPC Lasso has a pre-determined level of $\lambda$ that bounds the number of expected false positives and becomes more conservative as the correlation between the noise regressors increases.

\begin{lemma}
\label{lemma:fpc_fp}
	Assume that a strict mutual incoherence assumption holds:  $E(\bX_i ^T \bX_j) = 0, \hspace{2mm} \forall i \in S \wedge j \in F$. Let $\bX_{-j}$ be the design matrix $\bX$ excluding feature $j\in F$, $\bbetah_{-j}$ be the FPC Lasso estimator of the coefficient vector $\bbeta_{-j}$ of $\bX_{-p}$, and $\hat{\beps}_{-j} = \by - f(\bX^T_{-j} \bbetah^{\text{FPC}}_{-j})$. If the approximation error of the rwSNS $\bX_j^T \hat{\beps}_{-j} /\|\hat{\beps}_{-j} \|_2$ (see Appendix \ref{sec:sns} for further details on the rwSNS) is conservative in the tails, the FPC Lasso has the following three properties.
\begin{enumerate}[(i)]
    \item For any $j\in F$,
    \begin{align}
		P( j \in \hat{S}_{\lambda} ) &\leq 2[1-\Phi(\lambda)] \label{eq:prob_single}
    \end{align}
	where $\Phi$ is the standard normal CDF. That is, the probability of an individual false positive can be bounded.
    \item The expected number of false positives, denoted as FP, can be bounded. That is
    \begin{align}
		\text{FP} = E(|\{j: j \in \hat{S}_{\lambda}, j\in F\}|) \leq 2p[1-\Phi(\lambda)] \label{eq:prob_global}
    \end{align}
    \item The bound in \eqref{eq:prob_global} becomes increasingly conservative as the correlation between the columns of $\bX_F$ grow.
\end{enumerate}
\end{lemma}

By inverting equation \eqref{eq:prob_global}, one can easily obtain $\lambda^*_{\text{FP}}$, a value of $\lambda$ that can achieve a pre-specified expected number of false positives:
\begin{align}
\lambda^*_{\text{FP}} &= \Phi^{-1}\Bigg(1- \frac{\text{FP}}{2p}\Bigg) \hspace{2mm} \longleftrightarrow  \hspace{2mm }E\Big[\{\hat{S}_{\lambda^*_{\text{FP}}}\}_{j \in F}\Big] \leq \text{FP} \label{eq:fp_mu} .
\end{align}

Interestingly, the bound derived in \eqref{eq:prob_global} comes solely from the statistical properties of a rwSNS which characterizes the gradient (e.g. KKT conditions) of the FPC Lasso for the null features (see Appendix \ref{proof:fpc_fp}). By recasting the level of $\lambda$ in terms of an expected false positive count, there is an additional benefit that the FPC Lasso does not need to be extensively tuned and can be easily used by researchers who are not data science experts. However users would need to consider the likelihood of the strict mutual incoherence condition holding in their dataset. In situations where there is likely to be significant confounding, such as genomics, the FPC Lasso could return an inflated number of false positives. Strategies to ameliorate this issue can include using orthogonalized predictors (such as a PCA regression) or explicitly conditioning on treatment variables that are expected to have a causal relationship (also known as a double machine learning strategy \cite{glm2016}).

\subsection{FPC Lasso algorithm}

Algorithms which are able to solve the classical Lasso problem \eqref{eq:lasso_classical} have been extensively developed and solvers like \texttt{glmnet} \cite{friedman2010,simon2011} use a combination of coordinate descent, warm-starts, active set approximation, and screening rules \cite{tibshirani2012} to obtain state-of-the art performance. Because \texttt{glmnet} favours problems with a large number of features and/or a sequence of $\lambda$ values, it is well suited to solving the FPC Lasso problem as the one-to-one mapping function with the classical Lasso cannot be determined \textit{a priori} and instead must be discovered analytically. In addition to standardization (which is almost always used in any regularized setting), we also apply a skewness adjustment procedure as the distributional convergence of a rwSNS to the standard normal is a function of the skewness (see Appendix \ref{sec:sns}).

\begin{algorithm}
	\caption{FPC Lasso Algorithm solver using \texttt{glmnet} \label{alg:fpc_glmnet}}
	\begin{algorithmic}
		\Require Design matrix $\bX$, response $\by$, family $f$, tolerance $\epsilon$, target $\lambda^{\text{FPC}}$, and search parameters $\alpha \in (0,1)$ and $K \in \mathbb{R}^+$
		\State Minimize skewness of features with log transform and apply standardization
		\State $\bX_j \gets \log(\bX_j + c^*), \text{ where } c^* = \arg\min_c \hspace{2mm} \text{skew}(\log(\bX_j+c))^2$ for $j=1,\dots,p$
		\State $\bX_j \gets [\bX_j - \text{mean}(\bX_j)] / \text{sd}(\bX_j)$
		\State Determine $\lambda^{\text{max}} = \max_j |\bX_j ^T \beps_0|$, where $\beps_0$ is the residuals from the intercept-only model
		\State Initialize $\tilde{\lambda}^{\text{FPC}}=0$
		\While{$(\lambda^{\text{FPC}} - \tilde{\lambda}^{\text{FPC}})^2 > \epsilon$}
			\State Run \texttt{glmnet}$(\bX,\by,f)$ on a sequence from $\{\alpha \lambda^{\text{max}}, \dots, \lambda^{\text{max}}\}=\{\lambda_1,\dots,\lambda_K\}$
			\State Determine residual for each $\lambda_k$
			\State $\beps_{\lambda_k} \gets \by - f(\bX \bbeta_{\lambda_k})$
			\State Calculate equivalent FPC Lasso regularization level
			\State $\lambda^{\text{FPC}}_k \gets \lambda_k / \| \beps_{\lambda_k} \|_2 $
			\State Find closest match
			\State $ \tilde{\lambda}^{\text{FPC}} \gets \lambda^{\text{FPC}}_{k^*}$, where $k^* = \arg \min_k \hspace{2mm} (\lambda^{\text{FPC}}_k - \lambda^{\text{FPC}})^2$
			\State Find $\lambda_k$ to bound $\lambda^{\text{FPC}}$ and reset 
			\State $\lambda^{\text{max}} \gets \lambda_{i^*}$, where $i^*=\arg \min_i \hspace{2mm} \{ \lambda^{\text{FPC}}_i - \lambda^{\text{FPC}} | \lambda^{\text{FPC}}_i - \lambda^{\text{FPC}} > 0 \} $
		\EndWhile
		
		\Return{$\bbeta_{\lambda_{k^*}}$}
	\end{algorithmic}
\end{algorithm}

Algorithm \ref{alg:fpc_glmnet} considers a sequence of $\lambda$ values for the classical Lasso problem, calculates their equivalent regularization level for the FPC Lasso problem, and repeats this process until converegnce. While a proximal gradient algorithm could be used to solve \eqref{eq:lasso_fpc} directly, in practical settings we found this option to be slower than using \texttt{glmnet} to solve a sequence of problems. The use of existing solvers also has the advantage for certain edge cases including identifying whether the solution vector is completely sparse and if condition \eqref{eq:reg_lambda} is violated.

\section{Simulations\label{sec:sims}}

We conduct simulation studies to demonstrate two important properties pertinent to our FPC Lasso estimator. First, we examine whether the distributional properties of a rwSNS closely align with the quantiles of a standard normal distribution under a variety of distributions for a relatively small sample size. In terms of the FPC Lasso, this is equivalent to seeing whether $\bX_p \hat{\beps}_{-p} / \|\hat{\beps}_{-p}\|_2$ is close to a standard normal. Second, we show that FPC Lasso bounds the average number of false positives included across a variety of distributions, feature space sizes, and pairwise correlation coefficients. 

We applied the FPC Lasso to three different simulated response classes: continuous (Gaussian error), binary (logistic link), and right-censored survival time (Exponential distribution), each with the appropriate link function (Cox's PH model was used for the survival data).\footnote{The survival data also had a random censoring process that led to 25\% censoring, on average.} A sample size of $n=100$ was fixed for three high-dimensional scenarios: $p=\{100,1000,10000\}$, where the columns of the design matrix were drawn independently and identically distributed (i.i.d.) from either a Gaussian, Binomial, or Exponential distribution and then log transformed to minimize the empirical skewness and standardized with the empirical mean and standard deviation. A sparse coefficient vector was used with a coefficient magnitude of one size on ($\beta_{0j}=1$) and $|S|=5$. In the Gaussian response case, the error term was drawn from a standard normal.

\begin{figure}
	\includegraphics[scale=0.6]{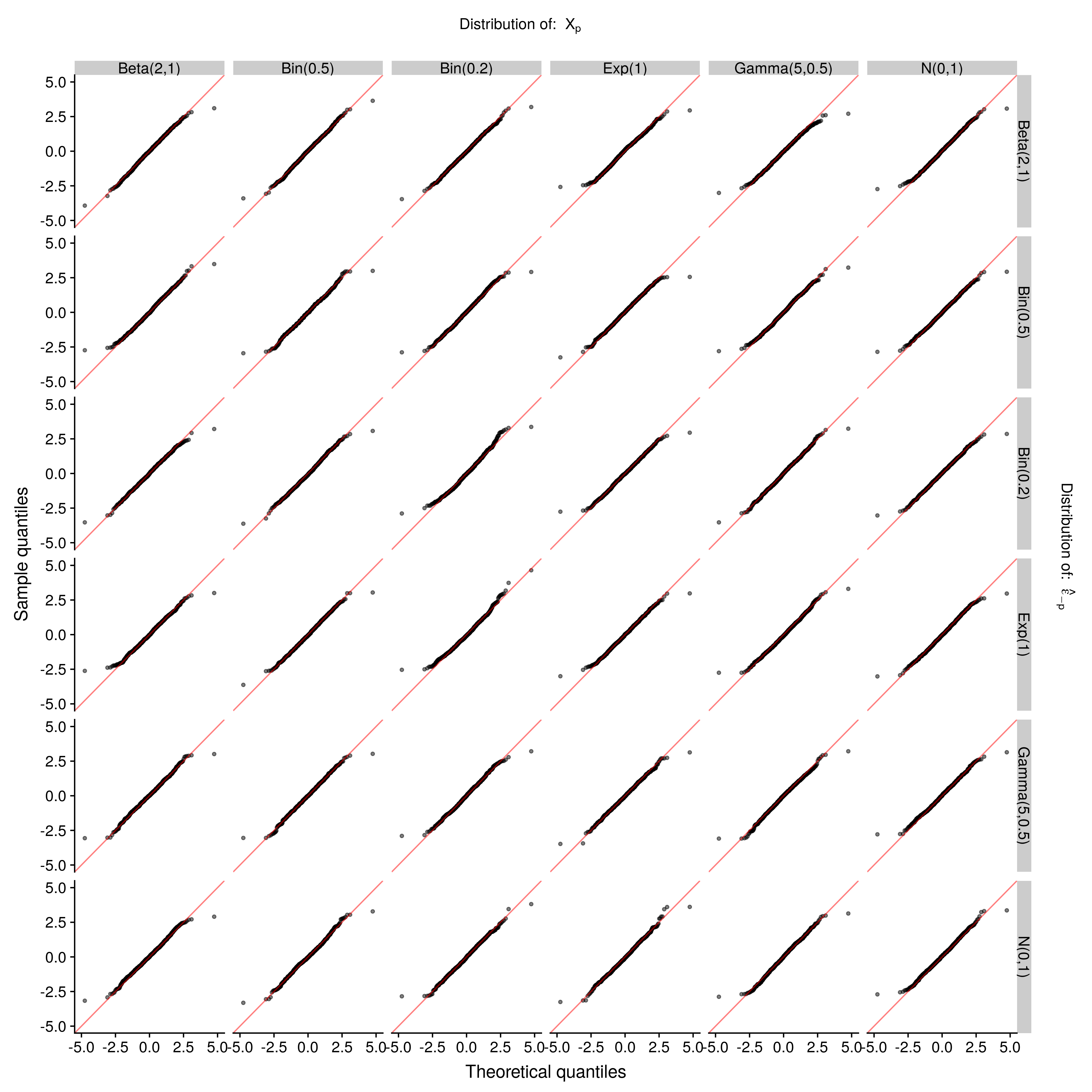}
	\caption{Q-Q Plot for rwSNS against standard normal with sample size 100 and 1000 replicates.\label{fig:qq_rwSNS}}
\end{figure}

Figure \ref{fig:qq_rwSNS} shows a panel of Q-Q plots of a standard normal distribution to the empirical quantiles of a rwSNS from a variety of distributions. Each plot is based on 1000 simulations with a sample size of $n=100$. As expected, the distributions between the empirical and theoretical quantiles are virtually identical. We also found that if the quantiles deviated in the tails, the rwSNS tended to have a conservative bias (i.e. shallow tails). 


\begin{figure}
	\caption{False positive control under an independent design matrix \label{fig:fp_indep}}
	\hspace*{-0.5cm}
	\includegraphics[scale=0.6]{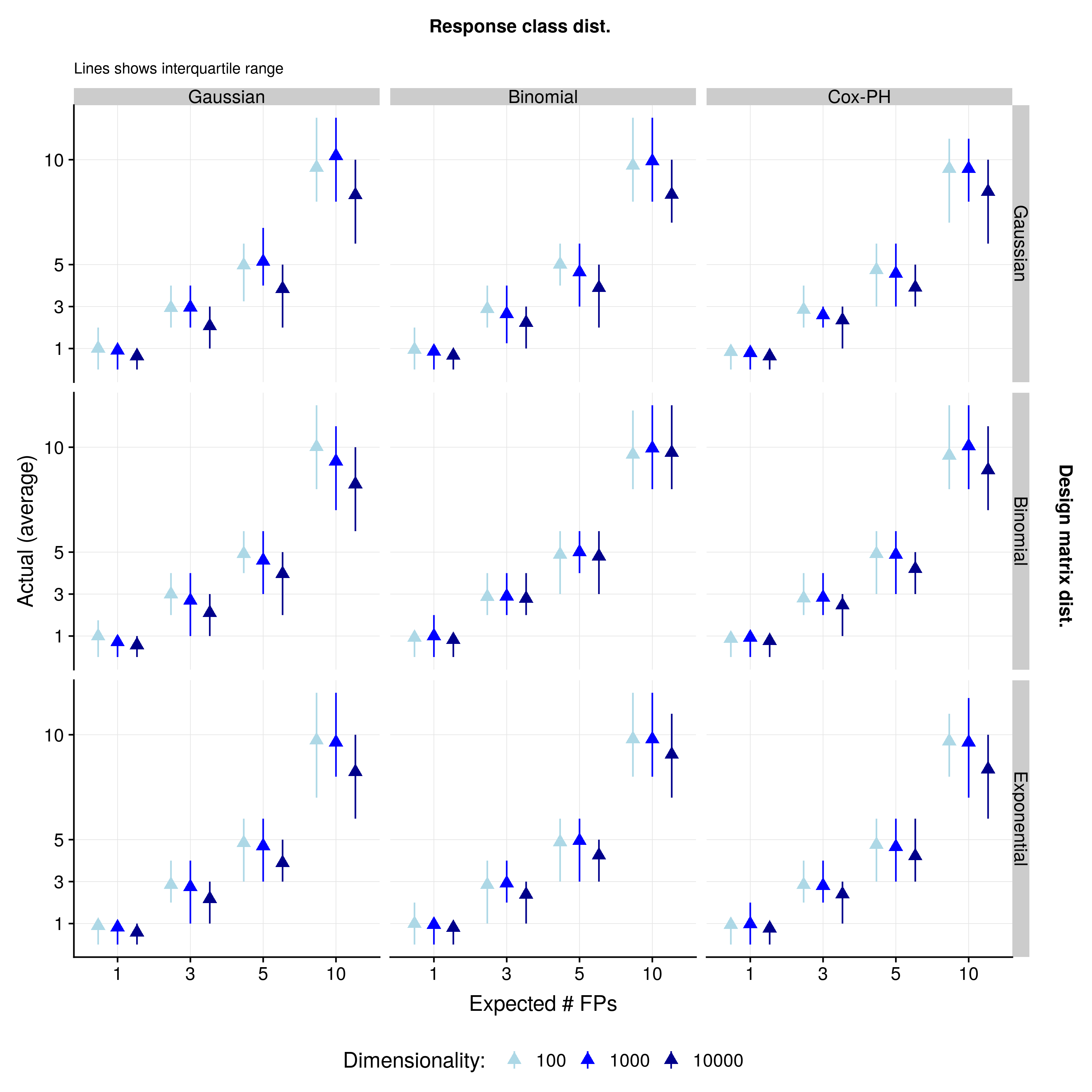}
	\floatfoot{$n=100$; $|S|=5$; $\beta_{0j}=1$; 250 simulations}
\end{figure}

The key property of our estimator is that its false positive control mechanism works well in finite samples, across GLM classes, and without making assumptions about the distribution of the individual columns of the design matrix (apart from the mutual incoherence condition) or the residuals. This is demonstrated in Figure \ref{fig:fp_indep}, which shows the average number of false positives selected by the FPC Lasso compared to the expected number based on a $\lambda$ choice from \eqref{eq:fp_mu}. As anticipated, the average actual number of false positives was bounded by our expected quantity in all cases. Notably the interquartile range around the number of false positives was also fairly tight. As the dimensionality $p$ of $\bX$ grew, the bound on the expected false positives became more conservative. We stress that this result could not have arisen because of approximation errors in the quantiles of the rwSNS since $n$ was fixed in all scenarios. Rather, this result stems from the growing number of columns that become coincidentally correlated from random chance alone as $p$ grows thereby reducing the effective dimensions of variation.

\begin{figure}
	\caption{True positive inclusion under an independent design matrix \label{fig:tp_indep}}
	\hspace*{-0.5cm}
	\includegraphics[scale=0.6]{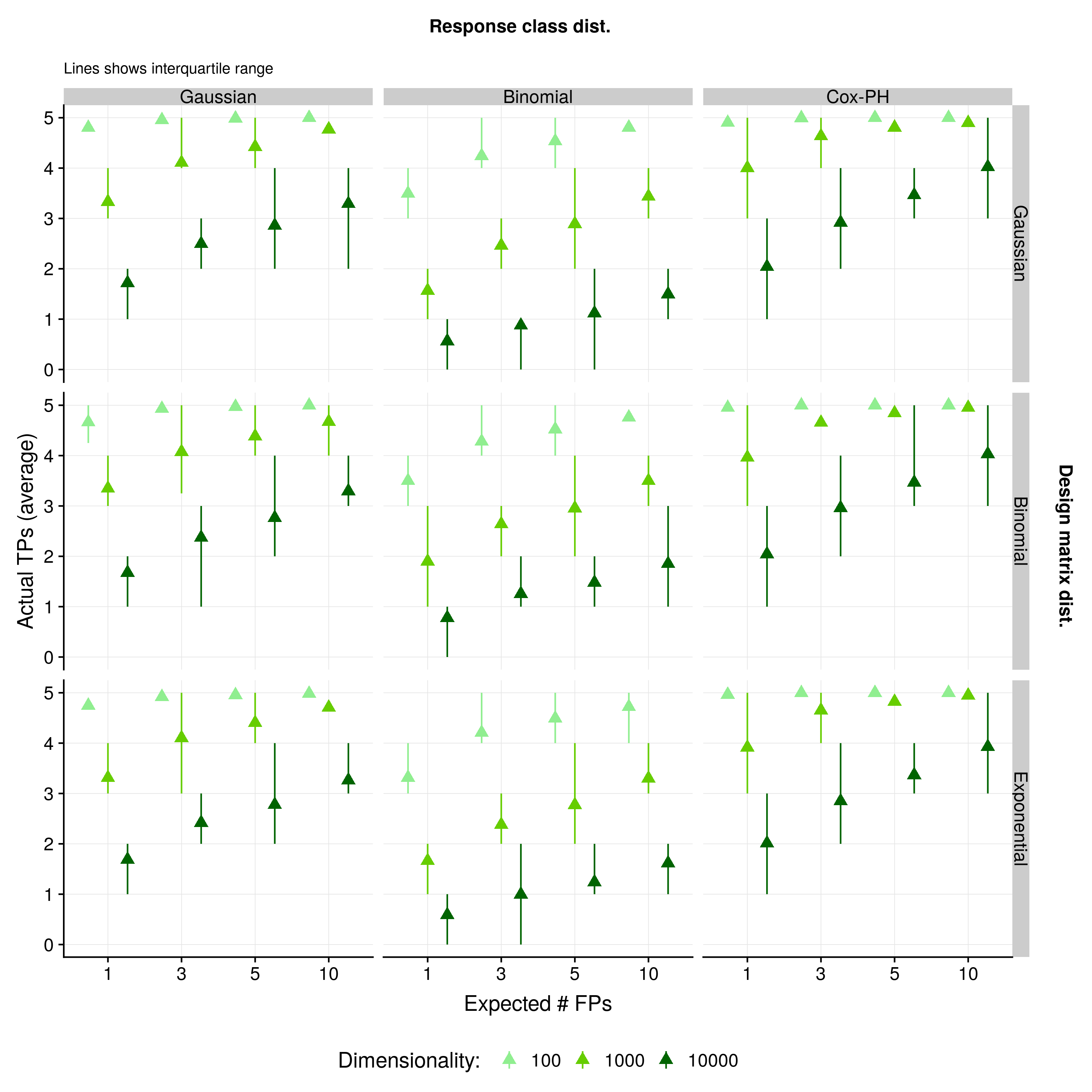}
	\floatfoot{$n=100$; $|S|=5$; $\beta_{0j}=1$; 250 simulations}
\end{figure}

Because the FPC Lasso has a fixed $\lambda$ for a given false positive bound, the expected number of true positives will be an increasing function of the coefficient magnitude size ($\beta_{0j}$), the signal to noise ratio (in the Gaussian case), and the ratio of $n/p$. Figure \ref{fig:tp_indep} highlights the fundamental trade-off between the number of false and true positives in the model. When $n \approx p$, perfect support recovery is almost feasible. However in the $p/n=10$ scenario, a bound of 10 false positives is required to recovery most of the five true positives. While the rate of true positive inclusion cannot be anticipated in advance, a post hoc estimate can be carried by dividing the pre-selected number of false positives by the size of the active set. 

\begin{figure}
	\caption{False positive control under a correlated design matrix \label{fig:fp_corr}}
	\hspace*{-0.5cm}
	\includegraphics[scale=0.6]{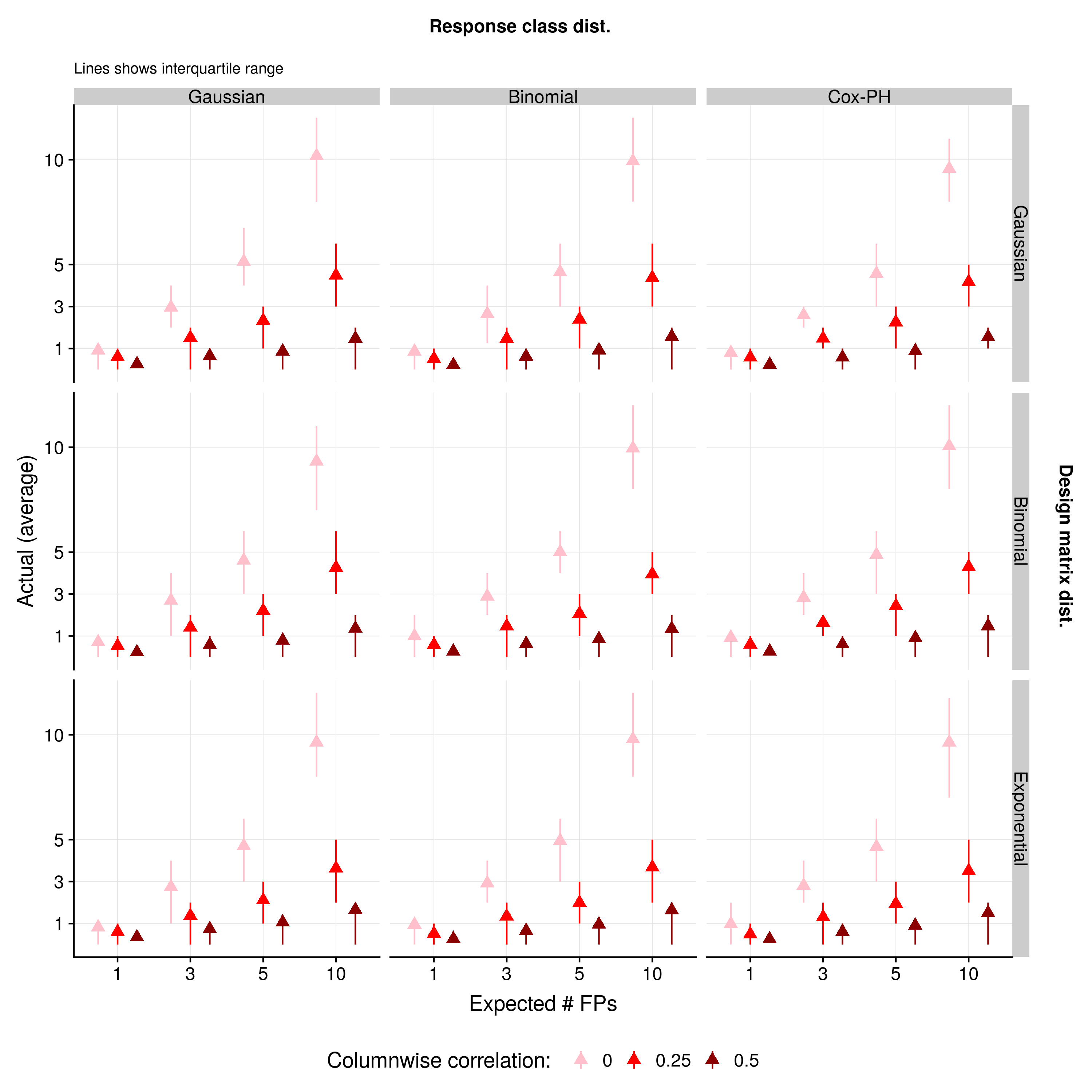}
	\floatfoot{$n=100$; $|S|=5$; $\beta_{0j}=1$; $p=1000$; 250 simulations; pairwise correlation over $\bX_F$}
\end{figure}

To further highlight how correlated columns in noise regressors lead to conservative bounds in the level of false positive inclusion, we repeated the above experiment by fixing $p=1000$ and $n=100$ and changing the pairwise correlation in the noise columns: $\text{cor}({[\bX_F]}_i,{[\bX_F]}_j)$. For example, when the pairwise correlation is 25\% and 50\%, a bound on ten false positives leads to an average of only around two and five respectively (see Figure \ref{fig:fp_corr}).   

A comparison between the performance of the FPC Lasso and other Lasso inference methods is challenging because there is not a direct relationship to the bounding of the expected number of false positives and other statistical measures including p-values and the false discovery \textit{rate} (because of Jensen's inequality the expectation of a ratio is not the same as a ratio of expectations). However we tried to approximate a fair comparison of the FPC Lasso to three competing inference approaches: (1) a sequential selection procedure \cite{gsell2016}, (2) the fixed-$\bX$ Knockoffs procedure \cite{barber2015}, and (3) exact post-selection inference \cite{lee2016,taylor2017}. We will refer to these approaches as G'Sell, Knockoffs, and selectiveInference, respectively, for the remainder of the paper. The \texttt{knockoff} and \texttt{selectiveInference} packages in \texttt{R} were used to carry out the comparisons. Because the G'Sell method requires a Gaussian model, and the Knockoff approach requires $n\geq 2p$ we repeated our simulation exercise from the previous section with a Gaussian response and a Gaussian design modifying $\beta_{0j}=0.5$ (so that perfect support recovery was unlikely), where $y_i \sim N(\sum_{j=1}^k \beta_{0j} x_{ij},\sigma^2)$. The sample size was set to $n=100$ and $p=100$, except for the Knockoff case which requires $n \geq 2p$ so $p=50$.

For the G'Sell approach, the FDR was set to 10\%. After the model returned the forward stepwise solution path that terminated when this rate was hit, we estimated the expected number of false positives as the length of the solution path multiplied by the FDR. The FPC Lasso was then fit with the regularization parameter bounding this expected false positve number. The Knockoffs method used the structured semidefinite program (SDP) algorithm to construct the design matrix and a cross-validated coefficient size difference as the symmetry statistic. Like G'Sell, the length of the solution vector was multiplid by the FDR rate of 10\% to obtain the expected false positive number. To carry out selectiveInference, a value of $\lambda = 0.5 \cdot \lambda^{\text{max}}$ was set and p-values were calculated. Assuming the p-values are uniform under the null of a zero effect size, we estimated the number of false positive as the mean number of p-values that were less than 10\%. Simulation draws in which the competing approaches returned an empty solution vector were ignored.

\begin{figure}
	\caption{True positive comparisons to other Lasso inference methods \label{fig:inference_comp}}
	\hspace*{-0.5cm}
	\includegraphics[scale=0.5]{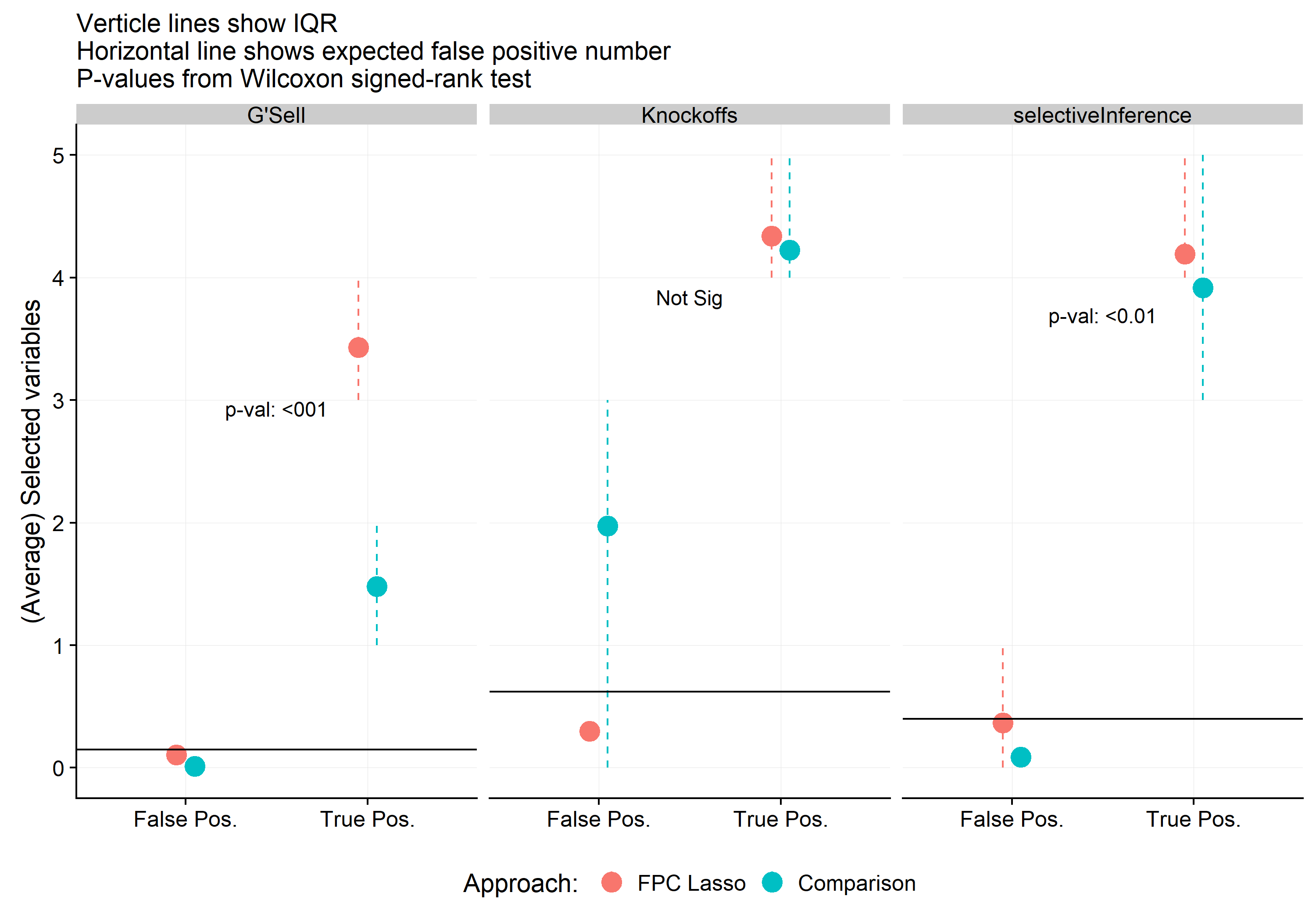}
	\floatfoot{$n=100$; $|S|=5$; $\beta_{0j}=0.5$; $p=50$ for Knockoffs and $p=100$ otherwise; 250 simulations}
\end{figure}

On average the FPC Lasso obtained a better true positive selection performance for a given expected false positive count compared to either G'Sell or selectiveInference (see Figure \ref{fig:inference_comp}). In the case of the Knockoffs, while there was no difference in the true positive selection rate, the number of false positives exceeded the anticipated amount suggesting that the power of the FPC Lasso would have been better at an equivalent false positive level. 

\section{Applied examples \label{sec:examples}}
We applied the proposed FPC Lasso algorithm in a double machine learning \cite{glm2016} procedure by selecting a parsimonious set of confounding variables that were related to both the time from palliative treatment to death (outcome) and the first-line ipilimumab vs first-line dacarbazine treatment for melanoma patients (a binary exposure). Details of the application can be found in \cite{drysdale2019}. Unlike more complicated models, the FPC Lasso can be easily used by clinical researchers as a bound on the expected number of false positives or false confounding variables is an easily interpretable measure. The linearity of the Lasso also adds to model transparency. 

\begin{figure}[!ht]
	\caption{FDR from the FPC Lasso in Cox's PH model applied to 31 datasets   \label{fig:real_data}}
	\hspace*{-0.5cm}
	\includegraphics[scale=0.6]{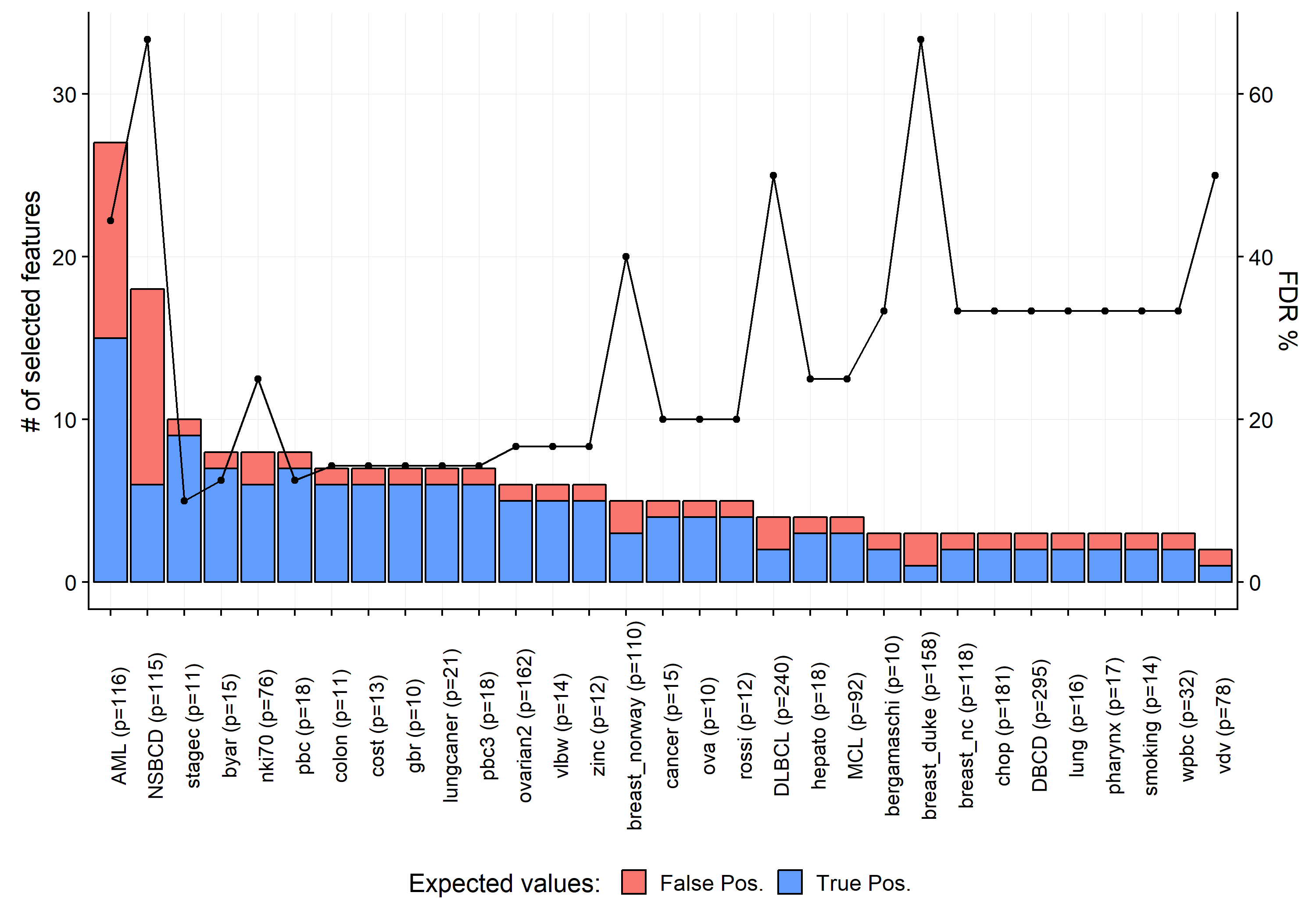}
	\floatfoot{FDR is approximate: ratio expected false positives to number of selected features}
\end{figure}

To examine the ability of the FPC Lasso to extract signal from noise on other real survival datasets, we considered 31 survival datasets that range from clinical datasets with 10 features to genomic datasets with 15K features, and Cox's PH model was employed to associate the features with the survival outcomes.\footnote{Most of our datasets were extracted from existing \texttt{R} packages that were found on CRAN including the \texttt{survival} package  \url{https://cran.r-project.org/web/views/Survival.html}.} To make the selection performance more comparable, the features were transformed to the $\min\{p,n\}$ principal components of each dataset. The level of $\lambda^{\text{FPC}}$ was set to select between $1,\dots,12$ expected false positives, and the level that had the lowest ratio of expected false positives to selected variables was chosen. The results are displayed in Figure~\ref{fig:real_data}. The genomic datasets (\texttt{AML}, \texttt{NSCBD}, \texttt{DLBCL}, etc) tended to have a higher ratio of false positives to overall selected features, suggesting the signal is more evenly distributed throughout the components of the design matrix. For relatively low-dimensional clinical datasets (\texttt{colon}, \texttt{gbr}, \texttt{stagec}, etc), the FPC Lasso tended to return most of the features. In all but two cases, the FDR was minimized by choosing an expected false positives number of one or two, suggesting that the FPC Lasso is able to obtain a relatively sparse model solution which aids in interpretability.

\section{Conclusion \label{sec:conclusion}}
As Lasso models become increasingly used as an inference technique in high dimensional settings, estimators whose false discovery properties hold in finite settings and across a range of GLM classes will become increasingly important.  The FPC Lasso can be used in many applied situations including high-throughput biology or covariate selection in treatment effect estimation for a variety of response tasks. In the prediction setting, it can be employed as a variable screener in which the variables in its support are reused for a secondary machine learning model. 

Unlike the selective inference approach or debiased estimators, the FPC Lasso does not have a differential inference strategy for the variables selected in its final solution vector (such as p-values). Instead the FPC Lasso uses an expected number of a false positives as a quantification of its uncertainty. The selected features in the returned model should be thought of as being highly associated with the response variable of interest, with the degree of plausible association tempered by the pre-chosen false positive bound. We believe that the FPC Lasso is particularly well suited for researchers who are not data science experts as the model can be implemented by picking a single hyperparameter. 

There are several drawbacks and limitations to the FPC Lasso. It is computationally more costly to estimate than the classical Lasso as it requires either solving part of the Lasso solution path to analytically determine the regularization parameter mapping or implementing an invex function optimization algorithm without the use of speedups like acceleration or backtracking line search. While we found the FPC Lasso's regularity condition ensuring a one-to-one mapping to be robust over the 151 datasets we tested it on (see Appendix \ref{proof:fpc_2prop}), this condition could be systematically violated on other datasets which would imply the presence of multiple solutions. The quantification of the expected number of false positives is also a measure of uncertainty that is not commonly used in statistics (marginal p-values along with false discovery and familywise error rates being the common metrics). Lastly the FPC Lasso suffers from a reliance on a strict mutual incoherence condition that may be acceptable in cases where the treatment variables are known in advance (so that possible confounders can be made orthogonal to these components) but could well lead to an inflated number of false positives otherwise. 

Future work for the FPC Lasso could include whether preconditioning the design matrix \cite{jia2015} could help to remove the stringency of the mutual incoherence condition and whether it can be extended to other sparse penalty approaches (such as the group Lasso). A more general study on the comparable performance of high-dimensional inference methods would also help to identify the strengths and weaknesses of different approaches. The results of this paper can be completely reproduced with freely accessible code and datasets available on \href{https://github.com/ErikinBC/fpclasso}{github}.

\clearpage


\bibliographystyle{plain}


\appendix
\section{Proof of Lemmas}

\subsection{Proof of Lemma \ref{lemma:fpc_2prop} \label{proof:fpc_2prop}}

Property (i) of the Lemma follows from the fact that if $\bbetah_{\lambda_1}^{\text{lasso}}$ satisfies the classical Lasso's KKT conditions $\bX^T\hat{\beps}_{\lambda_1}=\bgam \lambda_1$ from \eqref{eq:kkt_classical}, where $\hat{\beps}_{\lambda_1}=\by-f(\bX \bbetah_{\lambda_1}^{\text{lasso}})$ then it must also satisfy the FPC Lasso's KKT conditions $\bX^T\hat{\beps}_{\lambda_1}/\|\hat{\beps}_{\lambda_1}\|_2=\bgam \lambda_2$ for \eqref{eq:lasso_fpc} where $\lambda_2=\lambda_1 / \|\hat{\beps}_{\lambda_1}\|_2 $, making it at least a valid solution for the latter where $\bbetah_{\lambda_2}^{\text{FPC}} = \bbetah_{\lambda_1}^{\text{lasso}}$. If there is a strictly increasing and smooth relationship between $\lambda_2$ and $\lambda_1$, the rate of change of the residual vector must be bounded
\begin{align}
    \partial \lambda_2 / \partial \lambda_1 &> 0, \hspace{1mm} \lambda_1 \in (0,\lambda_{\text{max}}] \hspace{2mm} \longleftrightarrow \nonumber \\
    \frac{\|\hat{\beps}_{\lambda_1}\|_2 - \lambda_1 \frac{\partial}{\partial \lambda_1} \|\hat{\beps}_{\lambda_1}\|_2}{\|\hat{\beps}_{\lambda_1}\|_2^2} &>0  \hspace{2mm} \longleftrightarrow \nonumber  \\
    \|\hat{\beps}_{\lambda_1}\|_2 &> \lambda_1 \frac{\partial \|\hat{\beps}_{\lambda_1}\|_2}{\partial \lambda_1} \label{eq:ineq1}
\end{align}
which is the regularity condition seen in \eqref{eq:reg_lambda}. Previous work has demonstrated that any solution vector that satisfies the KKT conditions of the classical Lasso is a sufficient condition for optimality and that the classical Lasso has a unique solution in most cases \cite{tibshirani2013}.\footnote{The exceptions to this rule can occur when there are duplicate columns or for non-continuous features.} Under the situation where there are a unique sequence of $\bbetah^{\text{lasso}}_\lambda$ for $\lambda=(0,\lambda_{\text{max}}]$ for the classical Lasso, a proof by contradiction demonstrates why there cannot be another solution vector $\bbetah_{\lambda_1'}^{\text{lasso}}$ that also satisfies the KKT condition of the FPC Lasso. Without loss of generality consider the $j^{th}$ term which is in the active set of $[\bbetah_{\lambda_1}^{\text{lasso}}]_j > 0$, for $\bbetah_{\lambda_1'}^{\text{lasso}}$ to be a valid solution it holds that,

\begin{align*}
\frac{\bX_j^T \hat{\beps}_{\lambda_1'}}{\|\hat{\beps}_{\lambda_1'}\|_2} &= \frac{\lambda_1'}{\|\hat{\beps}_{\lambda_1'}\|_2} = \lambda_2 = \frac{\lambda_1}{\|\hat{\beps}_{\lambda_1}\|_2}.
\end{align*}

When $\lambda_1' \neq \lambda_1$ then $\lambda_2 = f(\lambda_1)$ cannot be strictly increasing function since $\lambda_2=f(\lambda_1)=f(\lambda_1')$ for $\lambda_1\neq \lambda_1'$. But since this violates the strictly increasing condition implied by \eqref{eq:ineq1}, $\bbetah_{\lambda_1'}^{\text{lasso}}$ cannot be a valid solution and satisfy the KKT conditions of the FPC Lasso confirming property (i) and (ii). 


Lastly, for property (iii), a function is said to be invex if and only if every stationary point is a global minimum. As we found empirical instances outside of the Gaussian case where the Hessian of the FPC Lasso's loss function was not positive semi-definite for all values of $\bbeta$, we know that its loss function is not convex. As the FPC Lasso solution is at least unique when it has a one-to-one mapping, it is therefore invex.

\begin{figure}[!ht]
	\caption{Relationship of $\lambda$ and $\lambda^{\text{FPC}}$ for regression datasets  \label{fig:lamseq_reg}}
	\hspace*{-0.5cm}
	\includegraphics[scale=0.4]{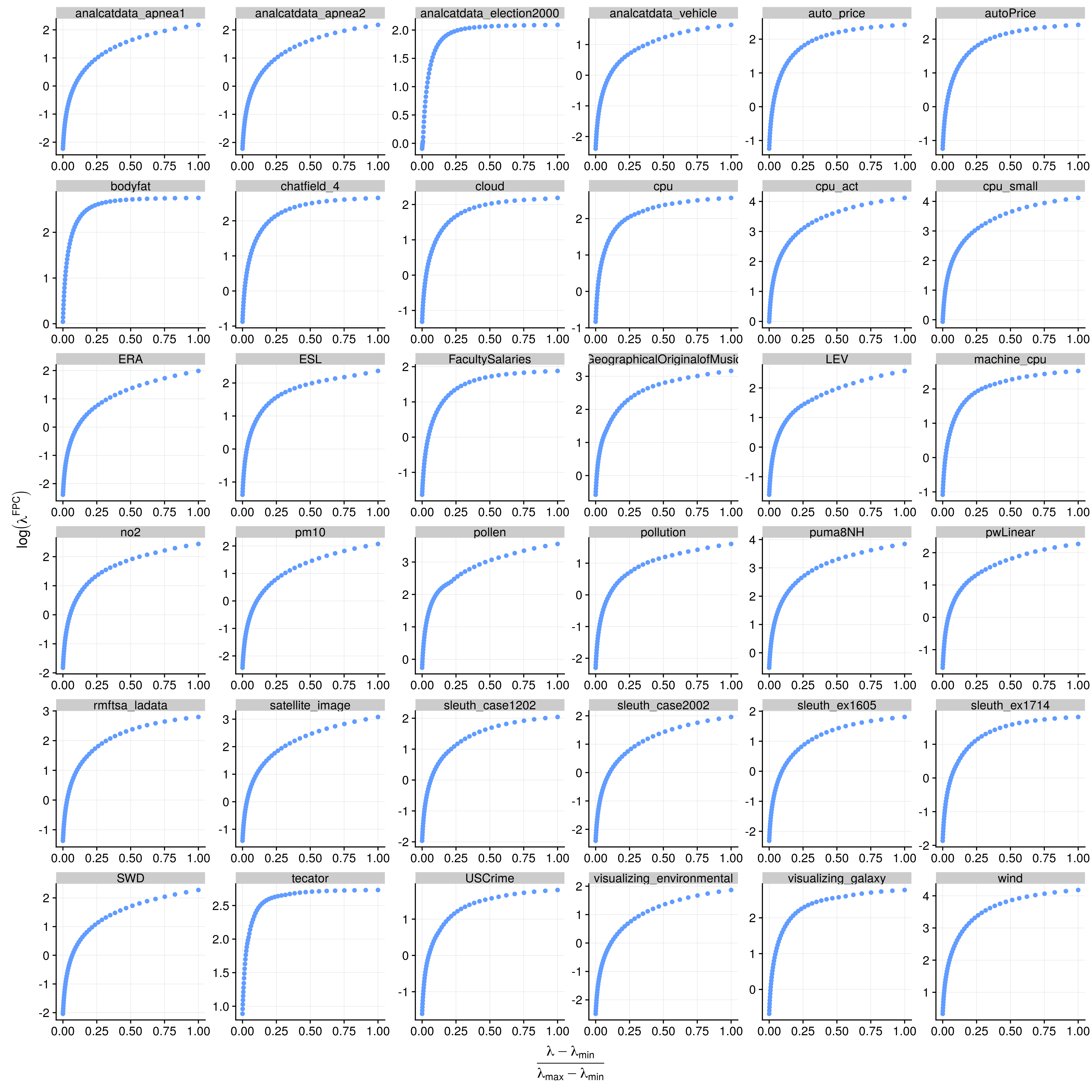}
	\floatfoot{$\lambda_{\text{max}}$ is the infimum of the completely sparse solution vector and $\lambda_{\text{min}}=0.01\cdot \lambda_{\text{max}}$}
\end{figure}
\begin{figure}[!ht]
	\caption{Relationship of $\lambda$ and $\lambda^{\text{FPC}}$ for classification datasets  \label{fig:lamseq_class}}
	\hspace*{-0.5cm}
	\includegraphics[scale=0.4]{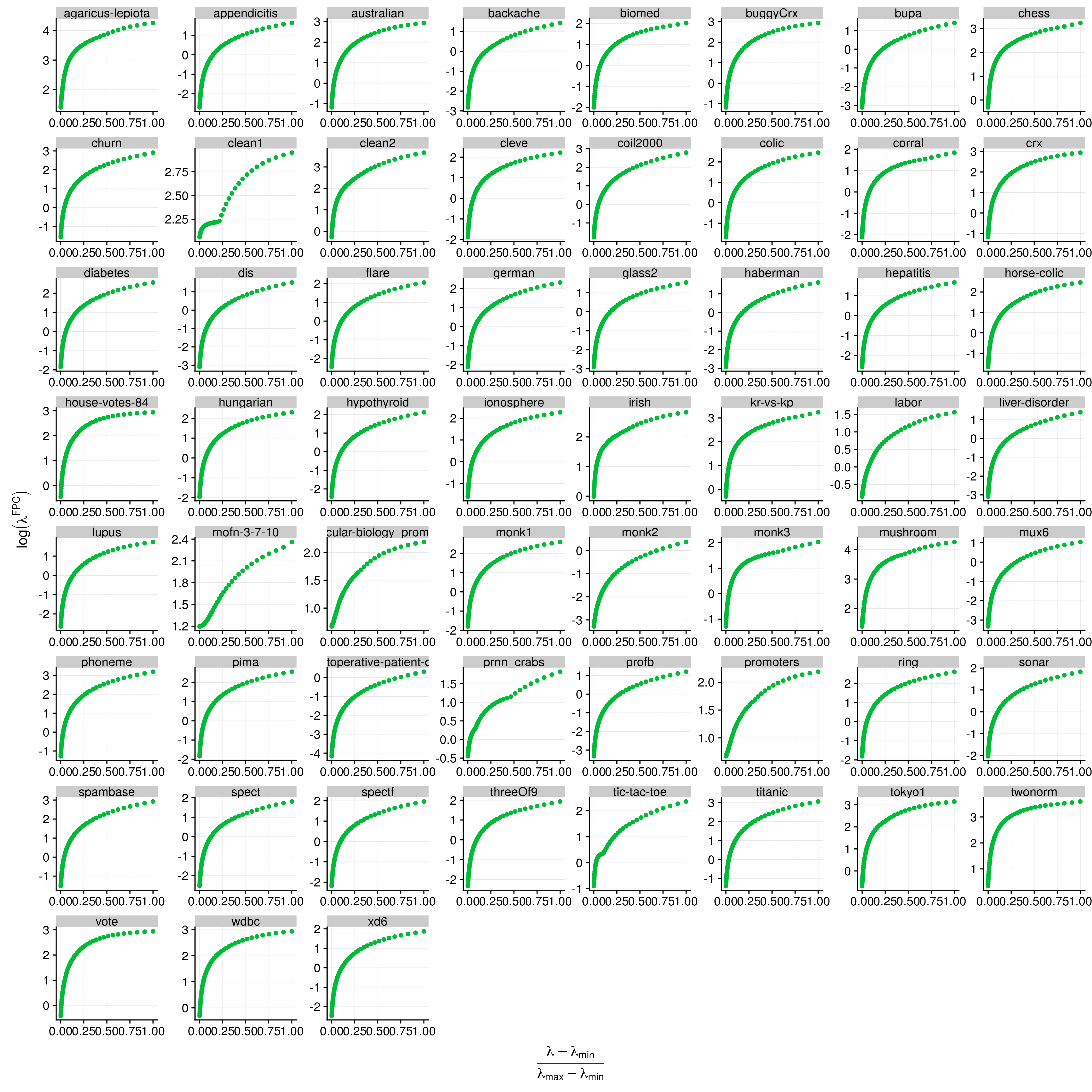}
	\floatfoot{$\lambda_{\text{max}}$ is the infimum of the completely sparse solution vector and $\lambda_{\text{min}}=0.01\cdot \lambda_{\text{max}}$}
\end{figure}
\begin{figure}[!ht]
	\caption{Relationship of $\lambda$ and $\lambda^{\text{FPC}}$ for survival datasets  \label{fig:lamseq_surv}}
	\hspace*{-0.5cm}
	\includegraphics[scale=0.4]{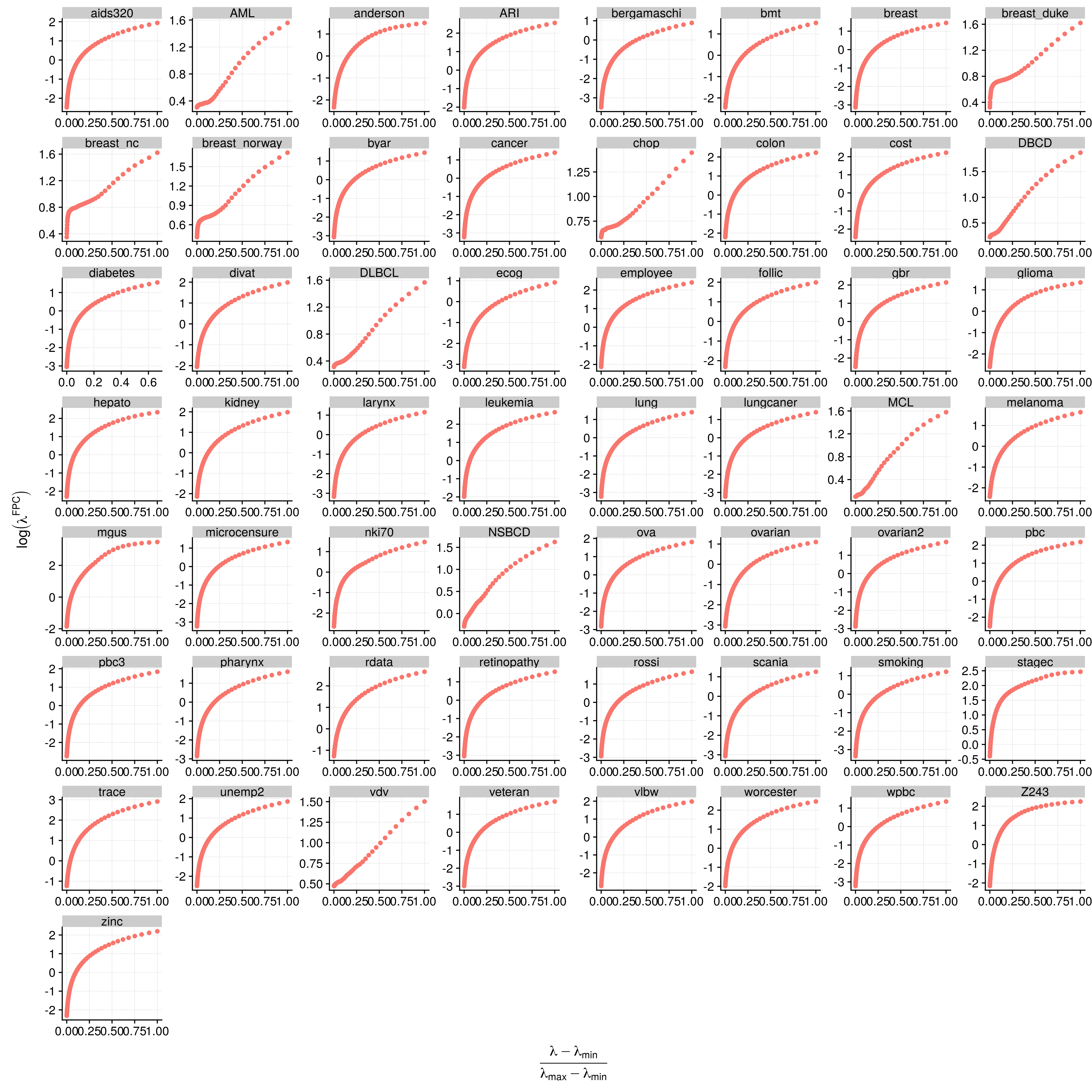}
	\floatfoot{$\lambda_{\text{max}}$ is the infimum of the completely sparse solution vector and $\lambda_{\text{min}}=0.01\cdot \lambda_{\text{max}}$}
\end{figure}

How likely is condition \eqref{eq:reg_lambda} to hold? We tested this condition over 151 datasets, 36 of which were regression problems, 59 were binary classification, and 57 were for right-censored survival data. While we did find instances where the inequality of \eqref{eq:reg_lambda} was violated, we stress that this was for only a single small interval along the solution path, giving us confidence that this technical condition is robust in practice. This can be seen in the monotonic relationship between the regularization parameter in the classical and FPC case that are shown in Figures \ref{fig:lamseq_reg}, \ref{fig:lamseq_class}, and \ref{fig:lamseq_surv}.

\subsection{Proof of Lemma \ref{lemma:fpc_fp} \label{proof:fpc_fp}}

Consider the FPC Lasso estimator fit to all the columns of the design matrix excluding the $p^{th}$ one, denoted as $\bX_{-p}$. Assume that regularity condition \eqref{eq:reg_lambda} holds so that there is a global minimizer. The FPC Lasso estimator of the coefficient vector of $\bX_{-p}$, denoted as $\bbeta_{-p}$, is
\begin{align}
\bbetah^{\text{FPC}}_{-p} &= \arg \min_{\bbeta_{-p}} \hspace{2mm}  \int_{\bm{-\infty}}^{\bm{\infty}} \nabla \ell(\bbeta;\bX_{-p}) d\bbeta + \lambda \| \bbeta_{-p} \|_1 \label{eq:lasso_notp}
\end{align}
We can determine whether the $p^{th}$ covariate will be part of active set in the full Lasso model \eqref{eq:lasso_fpc} by evaluating the inner product of its design matrix column, denoted as $\bX_p$, and the reduced model's residuals $\hat{\beps}_{-p} = \by - f(\bX^T_{-p} \bbetah^{\text{FPC}}_{-p})$. That is,
\begin{align*}
[\bbetah^{\text{FPC}}]_p &= \begin{cases}
|c| > 0 & \text{ if } \Big|\frac{\bX_p^T \hat{\beps}_{-p}}{\|\hat{\beps}_{-p} \|_2}\Big| > \lambda \\
0 & \text{ if } \Big|\frac{\bX_p^T \hat{\beps}_{-p}}{\|\hat{\beps}_{-p} \|_2}\Big| \leq \lambda.
\end{cases}
\end{align*}
The proof for both statements is as follows. If $|\bX_p^T \hat{\beps}_{-p} /\|\hat{\beps}_{-p} \|_2| \leq \lambda $ then the coefficient vector can be expanded with a zero entry and the KKT conditions will hold for $\bbetah^{\text{FPC}} = [\bbetah^{\text{FPC}}_{-p}; 0]$. Alternatively, if the absolute inner product exceeds $\lambda$ then the $p^{th}$ coefficient must be non-zero.\footnote{Although it is possible that one of the previously non-zero coefficients other than the $p^{th}$ one could become zero in the final solution.} A simple proof by contradiction reveals why this must be so. Suppose it was the case that $|\bX_p^T \hat{\beps}_{-p} /\|\hat{\beps}_{-p} \|_2| > \lambda $ and $[\bbetah^{\text{FPC}}]_p=0$, then the final model residual must be different $\hat{\beps}_{-p} \neq \hat{\beps}$ for the KKT conditions to hold and there must be a different solution vector for the first $p-1$ entries $[\bbetah^{\text{FPC}}]_{1:(p-1)} \neq \bbetah^{\text{FPC}}_{-p}$. But since $[\bbetah^{\text{FPC}}]_{1:(p-1)}$ satisfies the KKT conditions for \eqref{eq:lasso_notp}, this implies that there are at least two solutions to this problem, which is impossible as the Lasso has a unique solution and therefore $[\bbetah^{\text{FPC}}]_p\neq 0$.

As the $p^{th}$ column is a null variable, it is stochastically independent $\hat{\beps}_{-p}$ and hence $ \bX_p \hat{\beps}_{-p} / \|\hat{\beps}_{-p}\|_2$ will be characterized by a rwSNS process. That is,
\begin{align*}
P\Bigg(\Bigg|\frac{\bX_p^T \hat{\beps}_{-p}}{\|\hat{\beps}_{-p} \|_2}\Bigg| > \lambda \Bigg) &\approxeq 2[1 - \Phi(\lambda)].
\end{align*}
As noted in Appendix \ref{sec:sns}, a rwSNS converges in distribution to a standard normal distribution. In practical settings we have found the differences between the distributions to be negligible across a range of distributions, and where they are different, the error tends to be conservative: $\Phi_{\text{SNS}}(|\lambda|) < \Phi(|\lambda|)$ for larger absolute values of $\lambda$, where $\Phi_{\text{SNS}}$ is the CDF of the rwSNS and $\Phi$ is the CDF of a standard normal.

When the $p-k$ null columns of $\bX_F$ are independent of each other, the expected number of false positives will be equal to $E[\text{FP}] \approxeq 2p[1 - \Phi(\lambda)]$. However in almost all actual datasets, there is some correlation structure between the null features. It is a well documented phenomenon that the Lasso has a tendency to a select single representative feature from a group of correlated features. Indeed this was one the motivations for the development of the Elastic-Net model (see Theorm 1 from \cite{zue2005}). Unfortunately the statistical properties of the active set of the Lasso are notoriously difficult to characterize in finite samples with correlated features (in contrast the asymptotic properties are well characterized by the approximate message passing framework \cite{mousavi2017}). However when correlated features share the same coefficient sign it is easy to see why a single predictor could be selected. Consider a simple Lasso model with only two features $p=2$ that have been normalized with a mean-zero continuous response, then $\hat{\beta}_{2,\lambda} =T_\lambda(\bX_2^T(\by -\hat{\beta}_{1\lambda} \bX_1))= T_\lambda(\hat{\beta}_2^{\text{ols}} - \hat{\rho}(\hat{\beta}_1^{\text{ols}}-\lambda))$ where $T_\lambda$ is the soft-thresholding function, $\hat{\rho}$ is the empirical correlation coefficient between $\bX_1$ and $\bX_2$ and $\hat{\beta}_i^{\text{ols}}$ is marginal least squares coefficient value. Clearly as the absolute value of the correlation coefficient approaches one, the probability that the soft-thresholding function returns a zero increases. For this reason correlated features reduce the probability that the inner product with the existing residuals will exceed pre-specified threshold as the exiting correlated columns remove the direction of variation that the other columns have in common with the response. As simulation evidence shows in Section \ref{sec:sims} that the number of false positive features selected decreases significantly as the columnwise correlation in null features increases. 

\section{Self-normalized sums \label{sec:sns}}

Because randomly weighted self-normalized sum theory (rwSNS) plays such an important role in justifying the FPC Lasso's statistical properties we dedicate a small section to highlighting some of the distributional properties that we rely on that have been discovered in previous work \cite{hormann2013}. For a more general overview of self-normalized sum theory see \cite{shao2013}. We stress that remainder of this subsection is a restatement of properties that have been found in \cite{hormann2013} and we merely add commentary to relate it to the FPC Lasso where relevant. 

Let $\bA=\{A_n\}_{n\geq 1}$ and  $\{\bB_n\}_{n\geq 1}$ be two mutually independent random sequences. A randomly weighted self-normalized sum (rwSNS) is defined as
\begin{align}
\psi_n(\bB,\bA) &= \frac{\bB^T\bA}{\|\bA\|_2}.
\label{eq:rfsns}
\end{align}
A special case of the rwSNS is where $\psi_n(\boldsymbol{1},\bA)=\sum_{i=1}^{n}A_i / \sqrt{\sum_{i=1}^{n}A_i^2}$, in which the classical Student-t distribution can be written for example.  Another examples of a rwSNS is the empirical correlation coefficient. We now restate a key finding from \cite{hormann2013} that is directly related to the gradient of the FPC Lasso.\footnote{Lemma 2.1 in the original paper.} 

\begin{lemma} \label{lemma:sns}
	Let $\psi_n(\bB,\bA)$ be defined as in \eqref{eq:rfsns} and assume that $\bA$ and $\bB$ are mutually independent and i.i.d. which the following moments: $E(A_i)=0$, $\text{Var}(A_i) < \infty$, $E(B_i)=0$, $E(B_i^2)=1$, and $\xi_3= E |B_i^3| < \infty$. Denote $Z$ to be a standard normal random variable and $d_K(P,Q)$ and $d_W(P,Q)$ to be the Kolmogorov and Wasserstein distances  between probability measures $P$ and $Q$ with corresponding cumulative distribution functions $F$ and $G$ to be defined as follows
	\begin{align*}
		d_K(P,Q) &= \sup_x | F(x) - G(x) |, \hspace{2mm} x \in \mathbb{R} \\
	d_W(P,Q) &= \int_{0}^{1} |F^{-1}(t) - G^{-1}(t)| dt
	\end{align*}
Then
	\begin{align*}
	d_K(\psi_n(\bB,\bA),Z) &\leq 0.56 \cdot \xi_3 \cdot \delta\\
	d_W(\psi_n(\bB,\bA),Z) &\leq \xi_3 \cdot \delta,
	\end{align*}
	where
	\begin{align*}
	\delta=\sum_{i=1}^n E |\delta_{i,n}|^3, \hspace{2mm} \delta_{i,n}=A_i/\|\bA\|_2.
	\end{align*}
\end{lemma}

The important aspect of  Lemma \ref{lemma:sns} for our consideration is that if $\bB$ has a third moment equal to zero (i.e. no skew), then the Wasserstein and Kolmogorov distances between a rwSNS and the standard normal are zero and the quantiles of their distributions are identical. For this reason we seek to apply a skew-transformed and standardized predictor (to ensure $E(B_i^2)=1$) for the FPC model. In the more general case where $\xi_ 3\neq 0$, then slower convergence bounds have been derived.\footnote{See Theorem 2.3 and Example 2.2 in \cite{hormann2013}}

\begin{lemma} \label{lemma:rwsns_d}
	Assume that the tail behaviour of $\bA$ is known where $P(A_i^2 > x) \sim x^{-1}(\log x)^{-2}$, and $E(A_i^2) < \infty$, then $\forall n \geq 1$,
	\begin{align*}
	d_K(\psi_n(\bB,\bA),Z) &\leq A (\log n)^{-2},
	\end{align*}
	For a large enough constant $A$.
\end{lemma}

In practice we have found the rate of convergence of the rwSNS to the standard normal distribution to be very fast for a variety of distributions, and nowhere near as pessimistic as the logarithmic rates found in Lemma \ref{lemma:rwsns_d}. Nevertheless this Lemma gives an explicit rate of convergence to the normal law.

\end{document}